\documentclass[letterpaper, 10 pt, journal, twoside]{IEEEtran}

\IEEEoverridecommandlockouts                              

\usepackage{graphicx}
\usepackage{booktabs}
\usepackage{url}
\usepackage{amsmath}
\usepackage{lipsum}
\usepackage{graphicx}
\usepackage{tabularx}
\usepackage[dvipsnames,svgnames]{xcolor}
\usepackage{colortbl}
\usepackage{cite}
\usepackage{siunitx}
\makeatletter
\let\NAT@parse\undefined
\makeatother
\usepackage[colorlinks=true, citecolor=blue, linkcolor=blue, urlcolor=red]{hyperref}
\usepackage{cleveref}

\colorlet{CaptionColor}{gray!15}

\title{\LARGE \bf
CompSLAM: Complementary Hierarchical Multi-Modal Localization and Mapping for Robot Autonomy in Underground Environments
}

\author{Shehryar Khattak$^{*,1}$,  Timon Homberger$^{*,2}$, Lukas Bernreiter$^{3}$, Julian Nubert$^{4}$,\\Olov Andersson$^{2}$, Roland Siegwart$^{3}$, Kostas Alexis$^{5}$, Marco Hutter$^{4}$
\thanks{*Indicates equal contribution.}
\thanks{$^{1}$Author is employed the Jet Propulsion Lab, California Institute of Technology, USA. This work was done as an outside activity during author's previous affiliations$^{4,5}$.}
\thanks{$^{2}$Authors are with the Division of Robotics, Perception, and Learning, KTH Royal Institute of Technology, Sweden.}%
\thanks{$^{3}$Authors are with the Autonomous Systems Lab, ETH Z\"urich, Switzerland.}
\thanks{$^{4}$Authors are with the Robotic Systems Lab, ETH Z\"urich, Switzerland.}
\thanks{$^{5}$Author is with the Autonomous Robots Lab, NTNU, Trondheim, Norway.}
\thanks{This material is based upon work supported by the Defense Advanced Research Projects Agency (DARPA) under Agreement No. HR00111820045. The presented content and ideas are solely those of the authors.}
}

\begin{document}

\maketitle
\urlstyle{tt}
\thispagestyle{empty}
\pagestyle{empty}

\begin{abstract}
Robot autonomy in unknown, GPS-denied, and complex underground environments requires real-time, robust, and accurate onboard pose estimation and mapping for reliable operations. This becomes particularly challenging in perception-degraded subterranean conditions under harsh environmental factors, including darkness, dust, and geometrically self-similar structures. This paper details CompSLAM, a highly resilient and hierarchical multi-modal localization and mapping framework designed to address these challenges. Its flexible architecture achieves resilience through redundancy by leveraging the complementary nature of pose estimates derived from diverse sensor modalities. Developed during the DARPA Subterranean Challenge, CompSLAM was successfully deployed on all aerial, legged, and wheeled robots of Team Cerberus during their competition-winning final run. Furthermore, it has proven to be a reliable odometry and mapping solution in various subsequent projects, with extensions enabling multi-robot map sharing for marsupial robotic deployments and collaborative mapping. This paper also introduces a comprehensive dataset acquired by a manually teleoperated quadrupedal robot, covering a significant portion of the DARPA Subterranean Challenge finals course. This dataset evaluates CompSLAM's robustness to sensor degradations as the robot traverses 740 meters in an environment characterized by highly variable geometries and demanding lighting conditions. The CompSLAM code and the DARPA SubT Finals dataset are made publicly available for the benefit of the robotics community.\footnote{\label{foot:code}\url{https://github.com/leggedrobotics/compslam_subt}}~\looseness=-1
\end{abstract}

\section{Introduction}
This paper details CompSLAM, a complementary multi-modal robot localization and mapping framework designed for real-time, robust, and reliable robot operation, particularly in complex underground environments. The proposed method leverages visual, thermal, depth, inertial, and kinematic sensor inputs, employing a hierarchical coarse-to-fine fusion approach to generate accurate robot pose estimates and a consistent map. CompSLAM was developed and successfully deployed on a diverse fleet of legged, wheeled, and aerial robots of Team Cerberus during the DARPA Subterranean Challenge (2018-2021)~\cite{chung2023into}, culminating in its deployment on all robots of the team during their competition winning run of the final event. Beyond this competition, CompSLAM has also enabled a range of advanced robotic capabilities, including inter-robot map sharing~\cite{arm2023scientific}, ground-aerial robotic marsupial deployments~\cite{de2022marsupial}, multi-robot collaborative mapping~\cite{bernreiter2024framework}, and long-term construction operations~\cite{nubert2022graph,nubert2025holistic}.
Moreover, CompSLAM has served as a base system for two learning-based estimation methods, namely~\cite{nubert2021self}, where the regular feature-based scan-to-scan registration-based odometry has been replaced by a learned one, and~\cite{nubert2022learning} where a learning-based classifier is used to estimate localizability in degenerate self-symmetric environments.\looseness-1
%
\begin{figure}[t]
    \centering
    \includegraphics[width=\columnwidth]{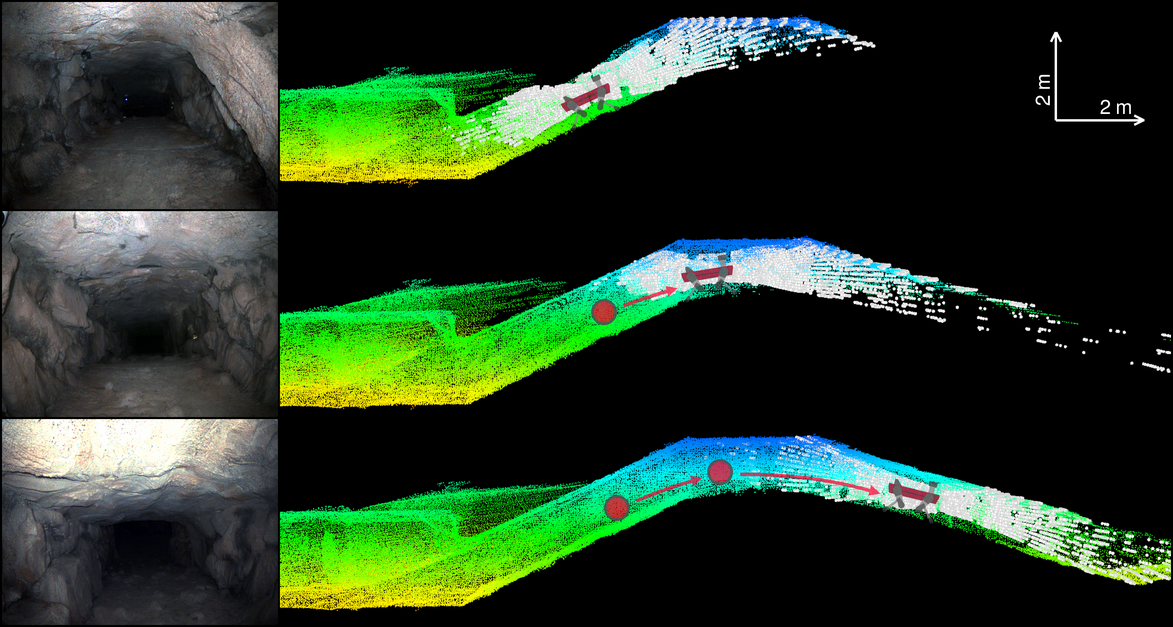} \caption{Narrow passage, traversed by the legged robot (location 4 in Figure~\ref{fig:dataset}). Left: Registered LiDAR pointcloud (white) and CompSLAM map (color), right: Corresponding images captured by the robot's onboard RGB camera.}
    \label{fig:narrow}
    \vspace{-2ex}
\end{figure}
%
\paragraph*{Paper Objective}
While prior publications have discussed aspects of CompSLAM within the context of the DARPA SubT Challenge~\cite{compslamICUAS,tranzatto2022cerberus,cerberusScience,cerberuswinsdarpa, ebadi2023present}, a comprehensive publication detailing the complete CompSLAM method and its broader applicability is currently missing. This paper addresses this gap with three primary objectives: \textit{\textbf{(a)}} to provide a unified and detailed description of the CompSLAM framework as implemented during the DARPA SubT Finals winning run and other developed capabilities demonstrated in subsequent projects; \textit{\textbf{(b)}} to release the CompSLAM source code to the public; and \textit{\textbf{(c)}} to share a comprehensive dataset collected using an ANYmal robot in the DARPA SubT Finals environment, covering most of the course, in contrast to the partial coverage achieved by the four robots during the actual competition. The authors aim to benefit the wider robotics research community by offering these three contributions.\footref{foot:code}\looseness-1


\vspace{-0.5ex}
\section{Related Work}
A large body of work has focused on developing robust and reliable estimation techniques over the past few years, particularly targeting real-world field deployments in challenging environments such as during the DARPA SubT competition.

\subsection{Sensor Fusion in Robotics}
The fusion of different sensor modalities to achieve better estimation, which in turn results in higher accuracy and increased robustness, has been a long-standing problem in robotics.
In modern robotic systems, IMU sensors are often at the core of the estimation problem, as done earlier in~\cite{mourikis2007multi} and~\cite{lynen2013robust}.
Historically, a large body of work has focused on filter-based estimators such as Extended Kalman Filters (EKF)~\cite{bloesch2017iterated}, Unscented Kalman Filters (UKF)~\cite{bloesch2013state} or two~\cite{bloesch2017two} or multi-state Kalman filters~\cite{mourikis2007multi,geneva2020openvins}.
More recently, there has been a push for optimization-based sensor fusion solutions in the field robotics community, due to their higher global accuracy and flexibility.
As an example, SuperOdometry~\cite{zhao2021super} deploys a central fusion factor graph, taking the output of two further factor graphs for visual-inertial odometry (VIO) and LiDAR-inertial odometry (LIO) as measurements, and feeding back the high rate fused estimate to the given submodules.
Moreover, Wisth et al. proposed a tightly fused formulation for legged robots in Vilens~\cite{wisth2022vilens}, taking as an input LiDAR, camera, leg kinematic and inertial measurements.
To operate outdoors in mixed environments, Nubert et al. proposed a dual factor graph formulation~\cite{nubert2022graph}, switching the optimization problem depending on the current environmental context, particularly focused towards the requirements in the domain of construction.
This formulation has then recently been generalized to arbitrary robotic systems and tasks in~\cite{nubert2025holistic} by offloading the switching logic into a single big factor graph optimization, allowing the handling of multiple measurements expressed with respect to multiple (different) reference frames.
CompSLAM LIO (this work) deploys a small factor graph in the LIO module to fuse-in inertial information to make the LiDAR odometry more robust.\looseness-1

\subsection{Lidar SLAM \& Odometry}
Lidar sensors have shown to perform well in challenging situations and environments over the past few years, in particularly during the DARPA SubT challenge~\cite{ebadi2023present}.
One of the first works to perform online LiDAR SLAM in practice was presented in LOAM~\cite{zhang2014loam}. While LOAM used point cloud features to run the system in real-time, more modern and faster compute units also allow the use of full ICP registration in a real-time system, as demonstrated in Open3D SLAM~\cite{jelavic2022open3d}.
Two systems that lay particular focus on the pointcloud undistortion by using continuous-time (CT) trajectory representation to model the trajectory are given in WildCat~\cite{ramezani2022wildcat} and DLIO~\cite{chen2023direct}, resulting in more sharp maps and robust registration even under agile motion.
Finally, Fast-LIO2~\cite{xu2022fast} constitutes one of the most widely used LiDAR-inertial odometry systems over the past few years due to its performance combined with low computational requirements due to its incremental KD tree formulation.
At the core, CompSLAM LIO uses feature extraction and registration in a similar fashion as LOAM, but additionally incorporates the prior from additional modules and fuses in inertial information, resulting in low compuational requirements while providing sufficient global and local accuracy.\looseness-1

\subsection{Estimation in Degenerate Environments}
For successfully deploying robotic systems in challenging environments, such as underground mines or tunnels, the used localization and mapping solutions have to handle both geometrically and visually degraded situations, in the form of self-symmetry~\cite{nubert2022learning}, darkness~\cite{khattak2019keyframe}, fog, dust or smoke~\cite{stanislas2021airborne}.\looseness-1

\paragraph{Explicit Handling of Degeneracy}
The first class of approaches explicitly handles degeneracy by either detecting it or estimating the uncertainty of the registration.
Cello-3D~\cite{landry2019cello} estimates the ICP uncertainty in real-world scenarios in order to more reliably fuse it with other modalities.
In contrast,~\cite{nubert2022learning} poses the degeneracy detection as a classification over \textit{localizability} for each of the 6 degrees-of-freedom in the body frame. 
X-ICP~\cite{tuna2023x} goes one step further by estimating the localizability along principle axes in the environment, also allowing the non-conservative detection across directions not aligned with the robot. Moreover, X-ICP also explicitly handles the degeneracy by adding constraints to the ICP optimization objective, which has been further investigated in~\cite{tuna2024informed}.
Finally, Ebadi et al.~\cite{ebadi2021dare} developed a degeneracy-aware LiDAR-based SLAM front-end to determine level of geometric degeneracy in an unknown subterranean environment.\looseness-1

\paragraph{Implicit Handling of Degeneracy}
Through a more implicit approach, Ebadi et al.~\cite{ebadi2020lamp} propose in LAMP a system consisting of a LiDAR-based frontend and an optimization-based backend specifically designed to handle outliers. Moreover, the fusion of multiple sensor modalities at the same time can help overcome degeneracy of single modalities, as e.g. done in SuperOdometry~\cite{zhao2021super} and Holistic Fusion~\cite{nubert2025holistic}, where the latter even explicitly models and estimates the drift of the used LiDAR odometry system.
In its original formulation, CompSLAM uses the degeneracy detection method of~\cite{zhang2016degeneracy}, but instead of simply remapping the solution to the observable directions, fallback estimates from the leg kinematics and VTIO are used to improve the odometry estimate (cf. \Cref{fig:overview}).\looseness-1

\section{Method}
\subsection{Overview}
\begin{figure}[t]
    \centering
    \includegraphics[width=\columnwidth]{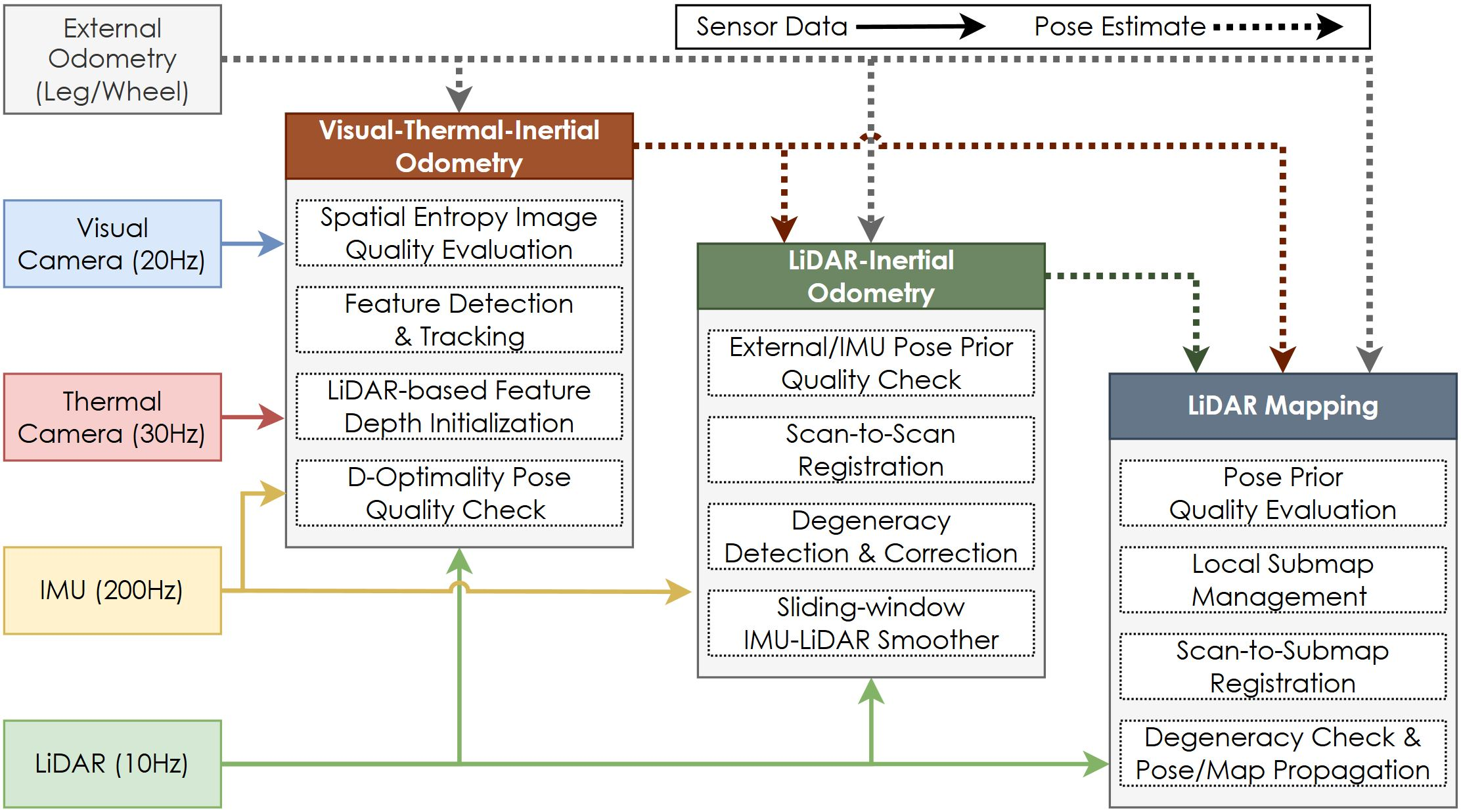}
    \caption{Overview figure of the proposed complementary multi-modal SLAM system showcasing different sensor inputs and the hierarchal refinement of robot odometry for reliable pose and map estimation.the different modules and passing of the estimated odometry as the initial guess of the next module.\looseness=-1}
\label{fig:overview}
\vspace{-3ex}
\end{figure}

The core design principle of CompSLAM is to provide a resilient robot pose estimation and mapping framework capable of delivering reliable robot pose and map updates at a consistent rate, even when the robot operates in perception-degraded environments. To be resilient, CompSLAM incorporates a suite of localization modules that can independently process combinations of multi-modal sensor data, including visual, thermal, depth, and inertial measurements, to generate individual pose estimates. This approach leverages the complementary nature of these sensor modalities under the assumption that environmental conditions are unlikely to degrade all sensor inputs simultaneously, thus ensuring operational robustness. 
Furthermore, to guarantee reliable pose and map updates at a desired frequency, CompSLAM employs a hierarchical, coarse-to-fine sensor fusion approach. For this, pose estimates generated by CompSLAM are fused with external estimates (e.g., kinematic odometry from legged robots). 
The key design idea here is to ensure resilience and reliability through redundancy. Suppose a localization module within the sensor fusion chain fails. In that case, it can be bypassed, and the pose estimate from the preceding step can be utilized by the subsequent pose estimation module and for map construction. This ensures the continuous availability of pose estimates and map updates crucial for supporting robot operation in complex and challenging environments. This approach is illustrated in Figure~\ref{fig:overview}, with detailed descriptions of each module provided in the subsequent sections.\looseness-1

\subsection{CompSLAM-VTIO}
The CompSLAM Visual Thermal Inertial Odometry (VTIO) module processes inertial measurements, at least one of or both visual and thermal images, and optionally LiDAR point clouds. To achieve real-time operation at high image update rates (visual: 20Hz, thermal: 30Hz), it employs an Extended Kalman Filter (EKF) to jointly estimate robot pose, IMU biases, and image feature states, drawing inspiration from the Iterated EKF (IEKF) approach of ROVIO~\cite{bloesch2017iterated}.\looseness-1
\paragraph{Image Quality Evaluation}
The maximum number of image features maintained in the state vector is fixed to bound computational cost. However, the number of incoming features from visual or thermal sources is dynamically managed based on feature availability and quality. Upon receiving an image, a spatial and temporal quality assessment is performed to select robust features. The image is divided into a grid of 4$\times$4 blocks, and the spatial entropy within each block is evaluated to assess gradient distribution. Blocks with spatial entropy exceeding the image's median value and a predefined minimum threshold are marked as candidates for feature detection. Before feature detection, the temporal consistency of each candidate block is assessed by tracking its validity across two consecutive frames. Given the high frame rate and assuming small robot motion, corresponding blocks are directly compared without warping to reduce computational overhead, relying on the assumption that the large block size ensures sufficient overlap for a reliable validity check. The quality assessment is designed to prioritize detecting image features with a higher probability of successful tracking, reducing the computational cost associated with frequent feature detection and patch extraction. This is achieved through two primary mechanisms: 
\textit{First}, it aims to avoid detecting features in image regions lacking sufficient gradient information necessary for reliable tracking. This is crucial because feature detection typically employs a significantly smaller kernel size than the patch size used for tracking. Localized image gradients might spuriously trigger feature detection in degraded visual environments such as darkness or fog. Yet, the corresponding larger extracted patch, centered on the detected pixel, may contain insufficient textural information for effective tracking. 
\textit{Second}, the quality assessment seeks to prevent feature detection in image areas exhibiting temporally inconsistent spatial entropy; for instance, on an aerial vehicle operating in darkness with onboard illumination, dust particles close to the camera can be brightly illuminated, covering substantial image areas in one frame but disappearing in the subsequent frame~\cite{khattak2019visual}. Detecting features on such transient phenomena would lead to unreliable tracking and negatively impact the overall pose estimation quality.

\begin{figure}[t]
    \centering
    \includegraphics[width=\columnwidth]{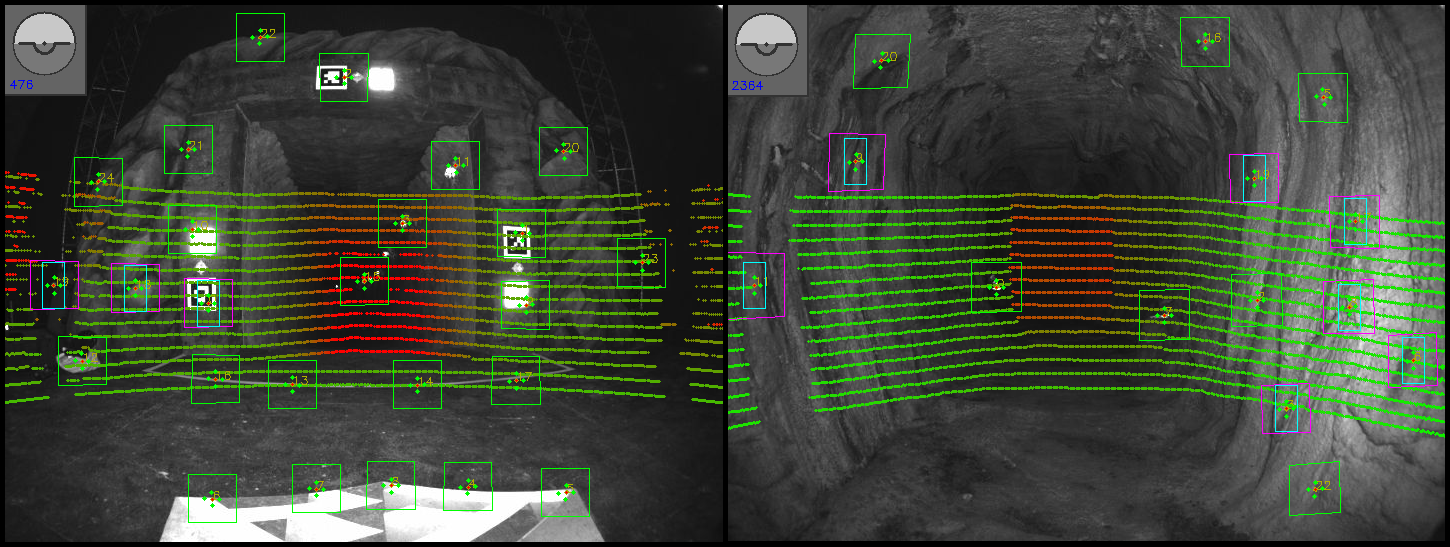}
    \caption{Tracked CompSLAM-VTIO image features and overlay of LiDAR points projected into the image frame. Features with purple-colored bounding boxes have been initialized using LiDAR depth.}
    \label{fig:combined-images}
    \vspace{-3ex}
\end{figure}

\paragraph{Feature Detection and Tracking}
For 8-bit visual images, features are detected using the FAST detector and tracked using multi-level patches, similar to ROVIO~\cite{bloesch2017iterated}. For thermal images, the full 16-bit radiometric information is utilized for both feature detection and patch tracking to prevent information loss associated with 8-bit normalization~\cite{khattak2019robust} and to enhance feature tracking robustness against noise accumulation in thermal imagery~\cite{khattak2019keyframe,khattak2020keyframe}.\looseness-1
\paragraph{State Estimation}
Image features are parameterized in the IEKF state vector using a bearing vector~\cite{trawny2006unified} and an inverse depth~\cite{civera2008inverse} parametrization. To maintain computational efficiency while integrating multi-modal image data, the thermal and visual cameras update their respective features independently, eliminating the need for cross-modality feature tracking. Consequently, VTIO can be viewed as a multi-modal N-monocular camera setup. While computationally efficient, monocular camera-inertial systems are susceptible to feature scale convergence issues, particularly during robot initialization before significant motion or when new features are added to the state vector.\looseness-1
\paragraph{Sparse Depth Initialization}
To mitigate this problem while leveraging an additional sensing modality for robustness, VTIO can optionally incorporate sparse LiDAR depth measurements during feature initialization or update steps. VTIO ingests undistorted LiDAR point clouds, transforms them to the timestamp of the current camera frame, and projects them onto either the visual or thermal image using the corresponding LiDAR-camera extrinsic and camera intrinsic calibration parameters. Before utilizing LiDAR depth for a feature point, depth verification is performed by considering all LiDAR points within a window centered around the feature's pixel coordinates. This verification ensures \textit{i)} a sufficient distribution of LiDAR points around the feature, requiring at least two points in each window quadrant, and \textit{ii)} local planarity of the scene around the feature by ensuring the depth standard deviation within the window is below a defined threshold. For newly initialized features with corresponding valid LiDAR depth, the median depth within the window is used to calculate the inverse depth and a small depth covariance is assigned. For features initialized without valid LiDAR depth, the median depth of all currently tracked features is used with a larger associated depth covariance. LIDAR measurements can sweep over features initialized without LiDAR depth as the robot moves. In such cases, the feature's depth covariance is checked.  Suppose it exceeds the small covariance value used for LiDAR-initialized features. In that case, the feature's depth is considered not yet well-converged and is reset using the new LiDAR depth and its associated covariance within the filter. This flexible approach allows for feature initialization without immediate depth information while ensuring that features related to reliable LiDAR depth contribute to faster scale convergence for the pose estimation filter (cf. \Cref{fig:combined-images}).
\paragraph{D-Optimality Pose Quality Check}
Before propagating the VTIO pose estimate to the subsequent CompSLAM localization module, a supplementary quality check is performed by monitoring the D-optimality metric~\cite{carrillo2012comparison} for the pose component of the estimator's covariance matrix. The D-optimality criterion provides a scalar measure of the covariance matrix's size (surface area), with larger values indicating greater uncertainty. If significant relative increases in the D-optimality metric are detected over a temporal window spanning multiple consecutive frames, suggesting a degradation in filter performance, the VTIO module can be re-initialized using the most recent pose estimate from the broader CompSLAM framework. To avoid scaling issues during the calculation of the D-optimality metric, the metric is tracked independently for the translational and rotational components of the pose covariance matrix~\cite{tuna2023x}. It is worth noting that while this additional system-level safety mechanism was implemented, it was not triggered during the actual DARPA SubT Finals event.\looseness-1

\subsection{CompSLAM-LIO}
The CompSLAM LiDAR Inertial Odometry (LIO) module processes inertial measurements and LiDAR point clouds. 
\paragraph{Overview}
The LIO module is structured into two sub-modules: LiDAR Odometry (LO) and LiDAR Mapping (LM), inspired by LOAM~\cite{zhang2014loam}, with the LO and LM modules operating at different rates, 10Hz and 5Hz, respectively.
\begin{figure}[b]
    \vspace{-3ex}
    \centering
    \includegraphics[width=\columnwidth]{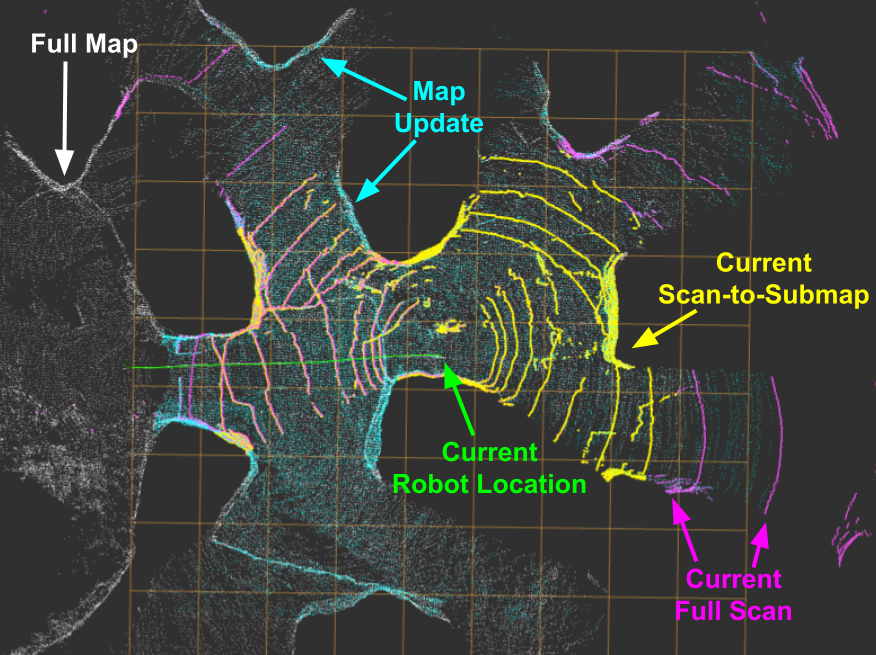}
    \caption{Highlighting different ranges of map used for scan-to-submap registration as compared to map update utilizing full range of the pointcloud.}
    \label{fig:submap}
\end{figure}
\begin{figure*}[t]
    \centering
    \includegraphics[width=\textwidth]{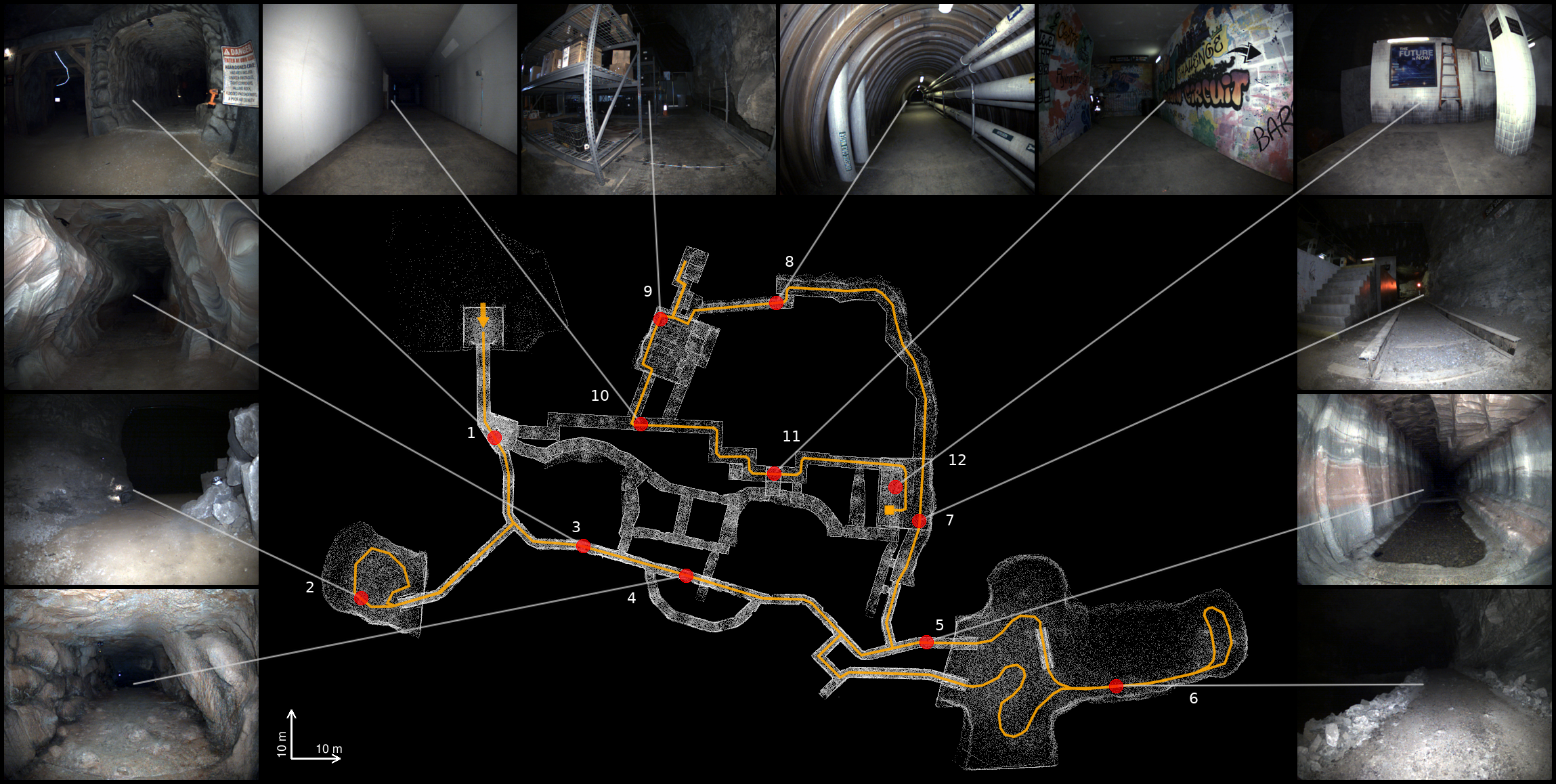}
    \caption{Sequence captured in the DARPA SubT final circuit environment, comprising tunnel, urban, and cave domains and featuring various geometries, such as self-similar tunnels, narrow passages, and large natural caves. The sample views are taken from the robot's onboard RGB stream. Corresponding locations and the approximate robot path are indicated on the ground truth point cloud map (GT: {\scriptsize\url{https://github.com/subtchallenge/systems_finals_ground_truth}}).}
    \label{fig:dataset}
    \vspace{-3ex}
\end{figure*}
The LO module performs scan-to-scan registration between the current and previous LiDAR point clouds. Like LOAM, line and plane features are extracted to efficiently register the point clouds by minimizing point-to-line and point-to-plane costs. 
It can also ingest multiple external pose estimates, prioritized according to user-defined order. For instance, on the ANYmal~\cite{hutter2016anymal} robot during the DARPA SubT Finals, the VTIO and leg odometry~\cite{bloesch2017two} poses were designated as primary and fallback pose estimates, respectively. 
Incoming external pose estimates are evaluated for their quality in order of priority to ensure that the provided relative translation and rotation estimates are within motion bounds allowed by the robot's controller. If a valid relative pose is found, it is used as an initial prior to the scan-to-scan registration step. If no valid external pose is available, IMU measurements between the timestamps of the two LiDAR point clouds are pre-integrated~\cite{forster2016manifold} to generate a motion prior.

\paragraph{Degeneracy Assessment}
To assess the quality of the point cloud registration, as it can become ill-conditioned in scenarios involving geometrically symmetric or self-similar environments, the eigenvalues of the approximate Hessian matrix underlying the optimization are monitored at each iteration~\cite{zhang2016degeneracy}. If any eigenvalue falls below a heuristic threshold, the registration is deemed degenerate, and the extracted point cloud features, along with the external pose estimates, are passed to the LM module. Analogous to the VTIO module, the eigenvalues corresponding to the translational and rotational components of the Hessian matrices are checked independently due to their differing scales~\cite{tuna2024informed}.\looseness-1

\paragraph{Factor Graph Smoothing}
To generate a smooth output from the LO module, the results of the scan-to-scan registrations are incorporated into a factor-graph-based fixed-lag smoother, implemented using GTSAM~\cite{dellaert2012factor}, along with IMU factors. The external relative poses are not added to the factor graph to avoid double-counting the same information, which is the same rationale for prioritizing them as priors for scan-to-scan registration over IMU-based priors when valid. Two additional factors are introduced into the factor graph during periods of no motion (\SI{0.5}{\second} or longer) to enhance the estimator's quality: \textit{First}, zero-velocity factors are added to improve IMU bias estimation. \textit{Second}, during these static periods, the global roll and pitch are directly estimated from the IMU accelerometer measurements and added to the factor graph to improve global alignment.

\paragraph{Local and Global Map Management}
The LiDAR Mapping (LM) module receives extracted point cloud features from the current scan and the refined pose estimate from the LO module to perform a scan-to-submap registration. Independently, the LM module also ingests external pose estimates, adhering to the same priority order as the LO module, to be used as a fallback if the scan-to-submap registration degenerates. Before utilizing these external fallback poses, the LM module independently validates their consistency, as its operational frequency may differ from that of the LO module. The LM module maintains an internal global map structured into cubic voxels with $10 \times 10 \times 10$\si{\meter} dimensions. Following each successful scan-to-submap registration, all registered points within the sensor's full range (e.g., \SI{100}{\meter} for a Velodyne VLP-16) are added to their corresponding map voxels, which are subsequently down-sampled to maintain a consistent map resolution. A submap KD-tree is constructed by concatenating the map voxels within a defined range of the robot's current estimated position. It is important to note that this submap range (e.g., \SI{50}{\meter} during DARPA SubT) is typically smaller than the full range of the current point cloud. This discrepancy arises from the observation that when performing scan-to-submap registration, distant map voxels may contain only sparse points, insufficient for reliably estimating the surface normal vectors required by the point-to-plane cost function. Therefore, scan-to-submap registration is performed within a shorter range. However, the registered point cloud is added to the global LM map at its full range. This allows distant map voxels to accumulate points over time as the robot navigates, ensuring that stable normal vectors can be extracted when these areas are later queried for submap creation. The reduced submap range also contributes to lower computational costs associated with building and searching the KD-tree. Additionally, a visibility check is performed for all points in the current scan to identify and incorporate only those map voxels that contain points from the current scan. An illustration of pointcloud range used for submap creation and map update are shown in Figure~\ref{fig:submap}. Similar to the LO module, an eigenvalue analysis of the optimization approximate Hessian is performed to detect potential ill-conditioning during the scan-to-submap registration. If the scan-to-submap registration is deemed degenerate, the LO-refined pose or an external fallback pose is used to register the current scan to the map, ensuring that an updated pose and map are consistently available for robot operation.\looseness-1

\subsection{Additional Modules}
Over the years, supplementary CompSLAM modules were developed to enhance robot operational capabilities. While not deployed during the DARPA SubT Finals run due to readiness or overall team strategy, they are included here for completeness and are part of the released code.

\paragraph{Inter-robot Map Sharing and Re-localization} 
CompSLAM facilitates map sharing between robots, enabling them to re-localize within each other's maps. This capability was demonstrated in the marsupial legged-aerial robot deployment detailed in~\cite{de2022marsupial}, where a legged robot served as a map server, exchanging point cloud feature maps and initial re-localization estimates with an aerial robot client.\looseness-1

\paragraph{Multi-robot Collaborative Mapping} 
For collaborative mapping of large-scale environments, robots can utilize the increased computational resources of a centralized mapping server to resolve inter-robot constraints. These constraints can then be broadcast to all participating robots, enabling them to align their internal estimates with the central mapping server. As detailed in~\cite{bernreiter2022collaborative, bernreiter2024framework}, CompSLAM's maplab-integrator module is designed to receive and integrate both unary (absolute) and relative pose constraints from the server into its pose graph. Moreover, the maplab-integrator can efficiently compare and remove outdated constraints while incorporating new ones whenever the broadcasted pose constraints are updated.\looseness-1

\paragraph{Learning-based Scan-to-Scan Registration and Localizability Estimation} 
CompSLAM can replace its traditional LiDAR Odometry (LO) module with the learning-based registration module DeLORA, as demonstrated in~\cite{nubert2021self}. Additionally, CompSLAM can integrate scan-to-scan localizability predictions from~\cite{nubert2022learning} to adapt the LO module's behavior in LiDAR-degenerate environments to improve robot operation robustness.\looseness-1
\vspace{-2ex}

\section{Dataset and Evaluation}
\subsection{Dataset}
\label{sec:dataset}

Alongside CompSLAM, this manuscript releases a dataset of ANYmal robot navigating the highly challenging environment in which the DARPA Subterranean Challenge final event took place.
Located in the Louisville Mega Cavern, the competition course was designed to represent three different types of underground environments: tunnel systems, urban underground, and natural caves. Each type made up a portion of the environment.

\paragraph{Data}
The data sequence was captured using a teleoperated ANYmal quadruped~\cite{hutter2016anymal}, equipped with the full sensor suite used by the team Cerberus~\cite{cerberuswinsdarpa}. The data contains LiDAR scans of a Velodyne VLP-16\footnote{Available as packets of raw or motion-compensated point clouds.} and grayscale camera sequences that are accurately time-synchronized with IMU readings using an Alphasense core module.\footnote{{\scriptsize\url{https://github.com/sevensense-robotics/core_research_manual}}} Two grayscale cameras are oriented along the robot's walking direction, allowing them to be used for stereo-visual-inertial-odometry methods. RGB image streams from 4 cameras are oriented in mutually orthogonal directions. These are recorded at a low frequency and serve as a reference. Furthermore, the sequence contains leg-kinematic-inertial odometry estimates. Intrinsic and extrinsic calibration of the camera setup and relative positions of all sensors are provided. Camera calibrations were derived using the Kalibr calibration toolbox~\cite{6696514}. Table~\ref{tab:sensor_info} serves as an overview of the available sensor data.\looseness-1


\begin{table}[t]
    \centering
    \caption{Dataset Measurement Types and Collection Rates}
    \vspace{-2ex}
    \begin{tabularx}{\columnwidth}{l|>{\centering\arraybackslash}X}
        \hline
        \rowcolor{CaptionColor}
        \textbf{Sensor Name} & \textbf{Frequency (Hz)} \\
        \hline
        Lidar Data (raw) & 10 \\
        Lidar PointCloud (undistorted) & 10 \\
        3× Camera Grayscale & 20 \\
        4× Camera RGB & 0.5 \\
        IMU Body & 400 \\
        IMU Alphasense Camera Setup & 200 \\
        Leg-kinematic Odometry & 20 \\
        \hline
    \end{tabularx}
    \vspace{-3ex}
    \label{tab:sensor_info}
\end{table}

\paragraph{Environment}
Figure~\ref{fig:dataset} shows the environment's characteristics and the route taken during the data collection.
Robot teleportation allowed for uninterrupted traversal of most of the course with a single robot.
In \SI{35}{\minute}, it walked a total distance of \SI{740}{\meter}.
Most of the course comprises narrow tunnels, with two sections consisting of spacious natural caves. One of the caves is \SI{60}{\meter} long, \SI{40}{\meter} wide, and \SI{11}{\meter} high. It has a multi-level layout with a traversable slope (location 6).
A section contains warehouse-like storage shelves with various objects, such as boxes and street cones (location 9). Furthermore, there are sections with reflective water puddles (location 5) and sections with featureless, highly self-similar tunnels (locations 10 and 11). Figure~\ref{fig:narrow} shows an especially tight constriction with adjacent narrow inclines in two directions and a ceiling height of around \SI{120}{\centi\meter}, which the robot navigated through (location 4).
Generally, there is dim lighting or no lighting at all. The environment's illumination was mainly caused by 4 LED-based light sources integrated into the robot's camera setup.\looseness-1

\begin{figure}[t]
    \includegraphics[width=\columnwidth]{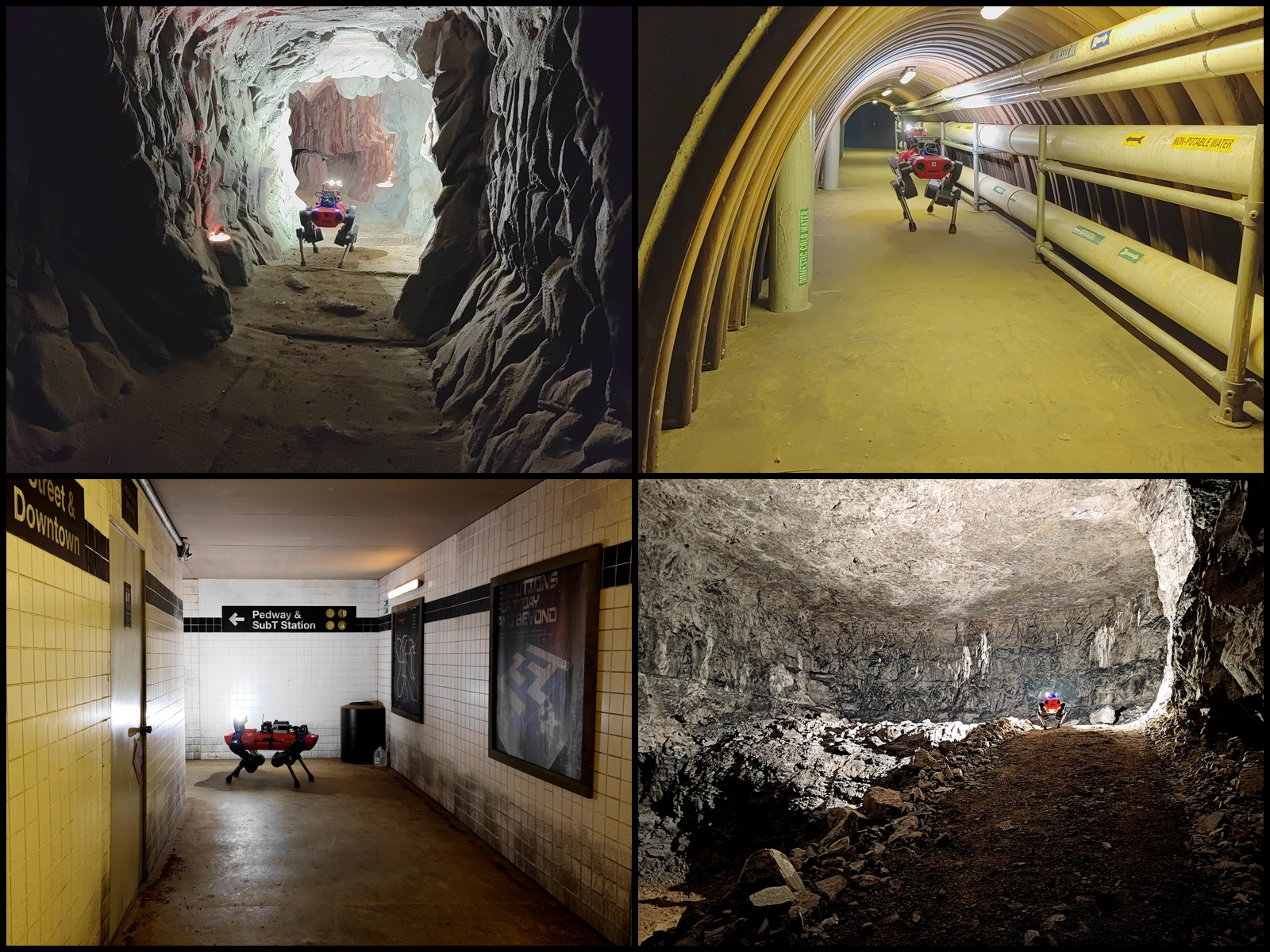}
    \caption{The ANYmal quadruped moving through the DARPA Subterranean challenge finals course during the dataset collection. Credit: Takahiro Miki.}
    \label{fig:datacollection_anymal}
    \vspace{-3ex}
\end{figure}

\paragraph{Challenges}
Challenges included, among others, missing LiDAR data due to spatial obstructions, poor geometrical constraints due to self-similar structures, and IMU saturation and motion blur in camera images due to abrupt motion on uneven terrain. Additionally, the camera images tend to feature skewed intensity distributions due to the demanding lighting conditions. Finally, moving people\footnote{The robot operators and journalists following the robot.} were periodically captured by the LiDAR. Figure~\ref{fig:datacollection_anymal} shows the ANYmal quadruped during the collection of this dataset.\looseness-1

\vspace{-2ex}
\subsection{Evaluation}
The released dataset, as described in Section~\ref{sec:dataset}, provides a challenging benchmark to evaluate SLAM approaches, not only in terms of accuracy but especially in terms of robustness to various sensor degradations. Therefore, it was used to evaluate CompSLAM.
For the evaluation, CompSLAM used ANYmal's leg-odometry as an external fallback pose estimate. Although all involved odometry sources incur drift over time, they remain robust throughout the entire data sequence. 

The estimated point cloud map $M_C$ is compared with the DARPA provide dense ground truth point cloud $M_{\text{GT}}$.
The point clouds are preprocessed by segmenting the first tunnel, including the three-way junction\footnote{Location 1 in Figure~\ref{fig:dataset} lies on the three-way junction.} from $M_C$ and $M_{\text{GT}}$ to produce $S_C$ and $S_{\text{GT}}$. Then, dense ICP alignment of $S_C$ to $S_{\text{GT}}$ is performed to get the relative transform $T_{\text{ICP}}$. $T_{\text{ICP}}$ is then applied to $M_{C}$, which yields $M_{C'}$.

\begin{figure}[t]
    \centering
    \includegraphics[width=1.0\columnwidth]{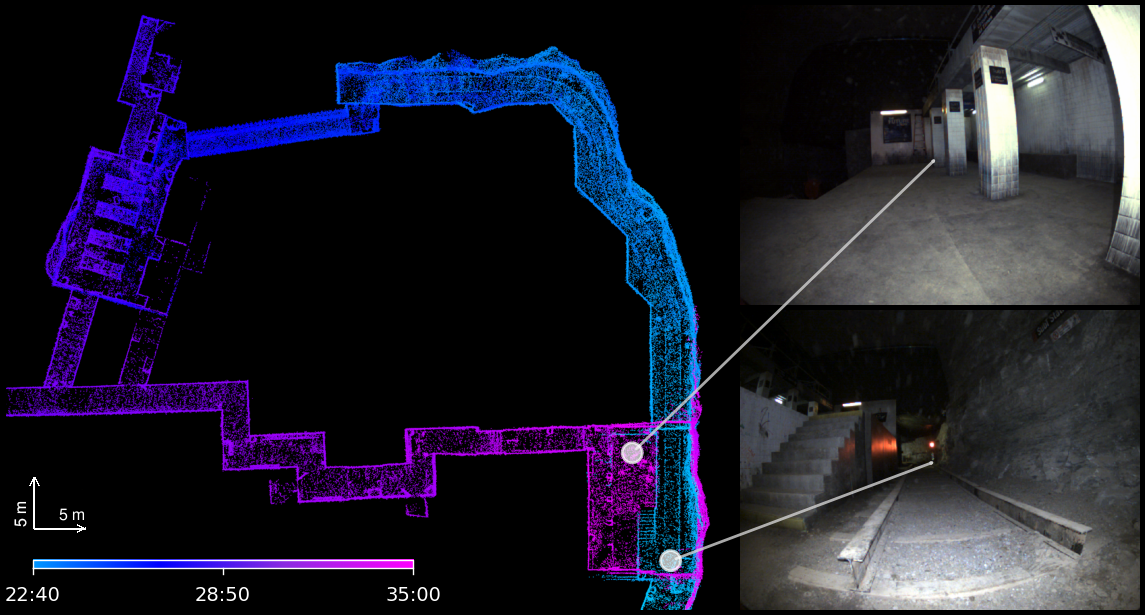} \caption{Top-down view of the estimated point cloud, with color reflecting the acquisition time of the LiDAR scans. 
    After passing the SubT-station along the rail tracks (\textbf{bottom right}), the robot returns to the station's elevated platform (\textbf{top right}). The depicted locations correspond to loc. 7 and 12 in \Cref{fig:dataset}.\looseness-1}
    \label{fig:loopclosure}
    \vspace{-3ex}
\end{figure}

\begin{figure}[b]
    \vspace{-2ex}
    \centering    \includegraphics[width=0.9\columnwidth]{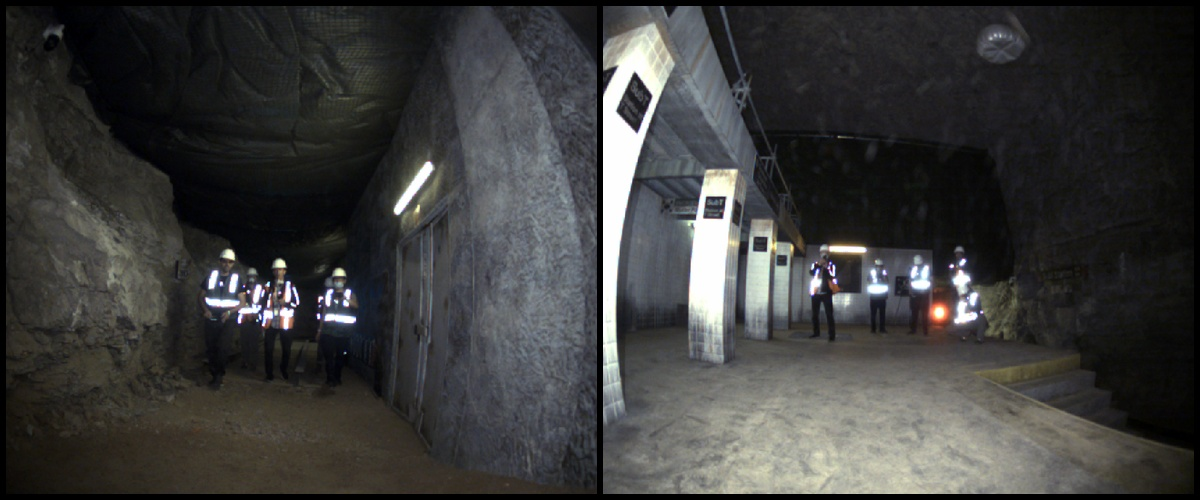} \caption{Moving persons, captured by the onboard RGB cameras. \textbf{Left:} Rail track section, between loc. 7 and 8 in \Cref{fig:dataset}. \textbf{Right:} SubT station, loc. 12.}
    \label{fig:people_moving}
\end{figure}

For a numerical comparison between $M_{C'}$ and $M_{\text{GT}}$, the recently proposed point cloud evaluation framework MapEval~\cite{hu2024mapeval} was employed.
It provides options for well-established point cloud evaluation metrics and introduces two novel metrics. The Average Wasserstein Distance (AWD) is used to assess robust global geometric accuracy, and the Spatial Consistency Score (SCS) is used for local consistency evaluation. These are derived with the help of spatial voxelization and per-voxel Gaussian statistics. Table~\ref{tab:awd_scs} presents results for different voxel side lengths. The results show an approximately linear increase in AWD with increasing voxel size, whereas SCS remains relatively stable. This observation is consistent with findings of a parameter sensitivity analysis that the authors report in~\cite{hu2024mapeval}.
\begin{table}[h]
    \vspace{-3ex}
    \centering
    \caption{MapEval Metrics: AWD and SCS}
    \vspace{-2ex}
   \begin{tabularx}{\columnwidth}{>{\columncolor{CaptionColor}}l| >{\centering\arraybackslash}X |>
   {\centering\arraybackslash}X |>
   {\centering\arraybackslash}X |>
   {\centering\arraybackslash}X |>
   {\centering\arraybackslash}X}
        \hline
        \textbf{Voxel Size (m)} & 1.0 & 2.0 & 3.0 & 4.0 & 5.0 \\
        \hline
        \textbf{AWD} & 0.221 & 0.554 & 0.892 & 1.219 & 1.612\\
        \textbf{SCS} & 0.491 & 0.516 & 0.530 & 0.583 & 0.640 \\
        \hline
    \end{tabularx}
    \label{tab:awd_scs}
\end{table}
\begin{figure}[t]
    \centering
    \includegraphics[width=\columnwidth]{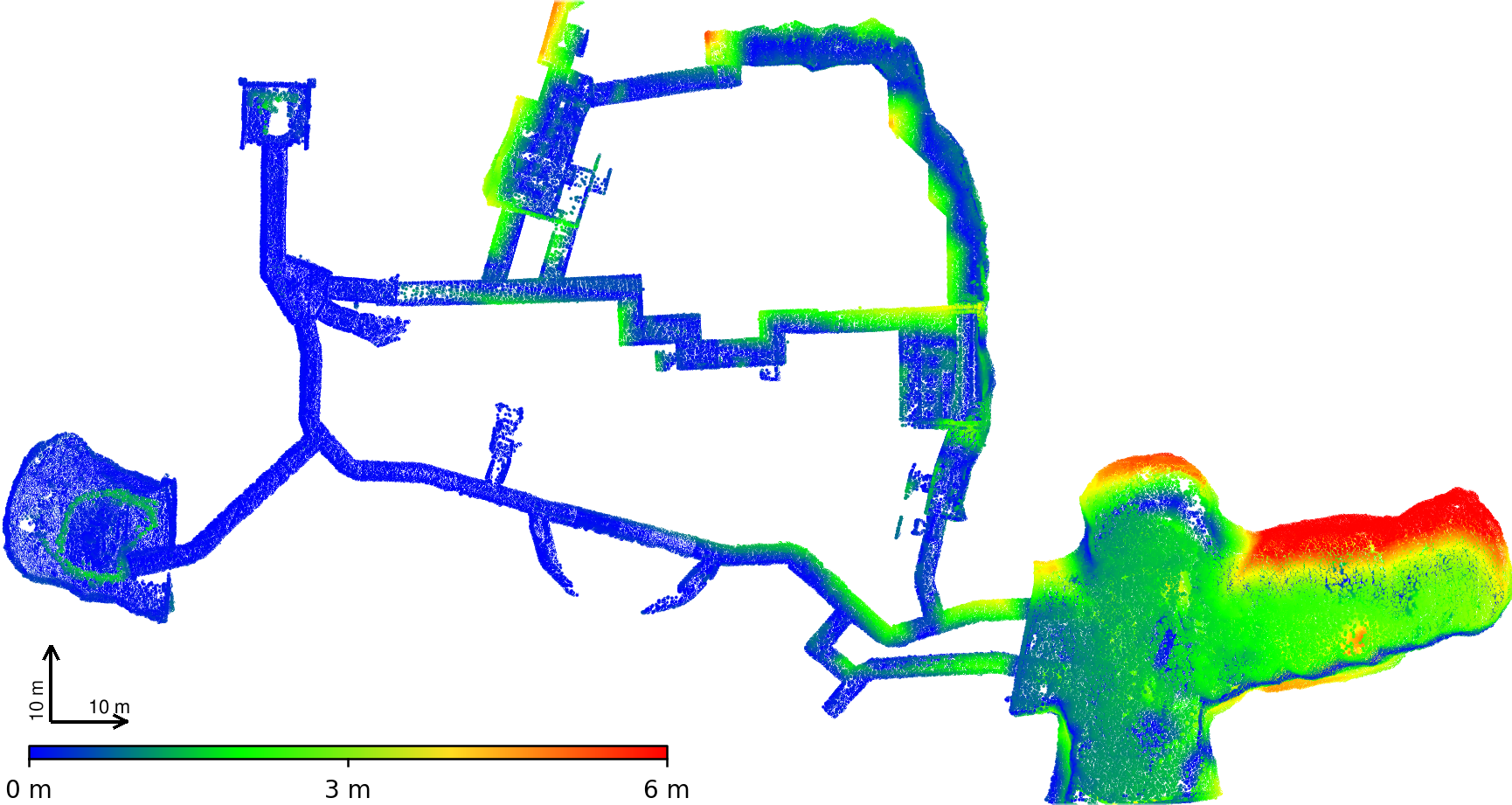}
    \caption{CompSLAM's estimated point cloud map. Colors show the euclidean distance of points with respect to the ground truth (truncation distance: 6.0 m).}    
    \label{fig:res_dist}
    \vspace{-2ex}
\end{figure}
Table~\ref{tab:acctable} shows the average Euclidean distance of estimated points to the closest ground truth point within a truncation distance.
Figure~\ref{fig:res_dist} shows the corresponding top-down view of $M_{C'}$ with colors reflecting the point-to-point Euclidean distances with respect to the ground truth.
\begin{table}[h]
    \vspace{-3ex}
    \centering
    \caption{Average euclidean distance of estimated points to GT}
    \vspace{-2ex}
    \begin{tabularx}{\columnwidth}{>{\columncolor{CaptionColor}}l|>
    {\centering\arraybackslash}X |>
    {\centering\arraybackslash}X |>
    {\centering\arraybackslash}X |>
    {\centering\arraybackslash}X |>
    {\centering\arraybackslash}X |>
    {\centering\arraybackslash}X}
        \hline
        \textbf{Truncation Dist. (m)} & 1.0 & 2.0 & 3.0 & 5.0 & 7.0 \\
        \hline
        \textbf{RMSE (m)} & 0.323 & 0.795 & 1.575 & 2.255 & 2.255 \\
        \hline
    \end{tabularx}
    \label{tab:acctable}
\end{table}
To quantify the accuracy of the estimated trajectory, a reference trajectory was generated, using offline ICP registration of the LiDAR point clouds with $M_{\text{GT}}$. Table~\ref{tab:rpe} shows the relative pose error between the estimated CompSLAM poses and the reference. The values were derived over trajectory segments of \SI{25}{\meter} length using the evo package~\cite{grupp2017evo}. The ICP reference trajectory and the method to produce it will be made available on the github.
\begin{table}[h]
    \vspace{-3ex}
    \centering
    \caption{Relative Pose Error (RPE) Components along the traejctory}
    \vspace{-2ex}
   \begin{tabularx}{\columnwidth}{l| >{\centering\arraybackslash}X |>{\centering\arraybackslash}X}
        \hline
        \rowcolor{CaptionColor}
        \textbf{Delta (m)} & \textbf{RPE translation (\%)} & \textbf{RPE rotation (deg/m)} \\
        \hline
        25.0 & 1.794 & 0.109 \\
        \hline
    \end{tabularx}
    \label{tab:rpe}
\end{table}
Qualitatively, as can be seen in Figure~\ref{fig:res_dist}, the point-to-point error increases after the robot traverses the elevated and narrow passage (Figure~\ref{fig:narrow}).
A misalignment affects the rest of the trajectory as it induces a global rotation error of $M_{C'}$ with respect to $M_{\text{GT}}$.
Figure~\ref{fig:loopclosure} shows a top-down visualization of a portion of the estimated point cloud $M_{C'}$. It is a loop that leads along the rails, past a subway station\footnote{This place was named SubT station and was designed to emulate a subway station as part of the urban section of the course.} through various types of tunnels and the warehouse section. It ends on the elevated station platform. The color gradient reflects the acquisition time of the LiDAR measurements as they get registered by CompSLAM. The robot traverses the \SI{220}{\meter} of the loop in approximately \SI{12}{\minute} and \SI{20}{\second}.
In Figure~\ref{fig:res_dist}, there are spurious points in the point cloud of the smaller of the natural caves. Lidar measurements of moving people cause points that do not correspond with the ground truth. This section does not have a noticeable impact on the robustness or accuracy of CompSLAM, as the correspondences of static structures appear unaltered. Figure~\ref{fig:people_moving} shows examples of people moving from the perspective of the robot's RGB cameras.

\section{Conclusion \& Lessons Learned}



This work provided a comprehensive overview of CompSLAM, a complementary, multi-modal SLAM system for resilient autonomy in highly demanding environments. Experiments on a data sequence representing the entire DARPA Subterranean Challenge finals course show the method's robustness to various sensor degradations. By releasing both the CompSLAM code and the dataset to the public, the authors aim to benefit the robotics research community.
While the method was designed and tested for underground exploration, its successful deployment in various projects, such as industrial excavation, demonstrates its wide array of use cases.

CompSLAM does not filter dynamic objects in its current version, which may affect its performance in dynamic scenes. Due to the presence of moving persons, the released dataset can be particularly useful for developments in dynamic SLAM.




\section*{Acknowledgments}
The authors are thankful to Marco Tranzatto, Samuel Zimmermann, Takahiro Miki, Lorenz Wellhausen, Gabriel Waibel for their assistance with ANYmal dataset collection.

\bibliographystyle{IEEEtran}
\bibliography{bibtex}

\end{document}